\documentclass[letterpaper]{article} 
\usepackage[]{aaai23}  
\usepackage{times}  
\usepackage{helvet}  
\usepackage{courier}  
\usepackage[hyphens]{url}  
\usepackage{graphicx} 
\def\UrlFont{\rm}  
\usepackage{natbib}  
\usepackage{caption} 
\usepackage{url}            
\usepackage{booktabs}       
\usepackage{amsfonts}       
\usepackage{nicefrac}       
\usepackage{microtype}      
\usepackage{xcolor}         
\usepackage[ruled,vlined]{algorithm2e}
\usepackage{graphicx}
\usepackage{pgf}   
\usepackage{amsmath, capt-of}

\nocopyright 

\title{Learning Adaptive Evolutionary Computation for Solving Multi-Objective Optimization Problems}
\author {
    R.H.M. Coppens\textsuperscript{\rm 1},
    R.V.J. Reijnen\textsuperscript{\rm 1},
    Y. Zhang\textsuperscript{\rm 1},
    L. Bliek\textsuperscript{\rm 1},
    B. Steenhuisen\textsuperscript{\rm 2}
}
\affiliations {
    \textsuperscript{\rm 1} Eindhoven University of Technology\\
    \textsuperscript{\rm 2} Nobleo Manufacturing\\
    remcocoppens@outlook.com, r.v.j.reijnen@tue.nl, yqzhang@tue.nl, l.bliek@tue.nl, berend.steenhuisen@nobleo.nl
}

\begin{document}

\maketitle

\begin{abstract}
Multi-objective evolutionary algorithms (MOEAs) are widely used to solve multi-objective optimization problems. The algorithms rely on setting appropriate parameters to find good solutions. However, this parameter tuning could be very computationally expensive in solving non-trial (combinatorial) optimization problems. This paper proposes a framework that integrates MOEAs with adaptive parameter control using Deep Reinforcement Learning (DRL). The DRL policy is trained to adaptively set the values that dictate the intensity and probability of mutation for solutions during optimization. We test the proposed approach with a simple benchmark problem and a real-world, complex warehouse design and control problem. The experimental results demonstrate the advantages of our method in terms of solution quality and computation time to reach good solutions. In addition, we show the learned policy is transferable, i.e., the policy trained on a simple benchmark problem can be directly applied to solve the complex warehouse optimization problem, effectively, without the need for retraining. 
 
\end{abstract}






\section{Introduction}\label{sec:INTRO}

In Combinatorial Optimization Problems (COPs), the goal is to identify high-quality solutions. 
Given that most COPs are NP-hard, solution approaches typically rely on heuristics to achieve good solutions in short computation time, compromising the optimality.  
Machine learning (ML) has recently been successfully leveraged for learning to solve classical COPs such as Travelling Salesman Problem (TSP) and other complex optimization problems  \cite{kool2018attention,kwon2020pomo,hottung2020learning,da2021learning}.  

In comparison to the amount of attention spend on single-objective COPs, much less work has been done in the ML community on solving multi-objective optimization problems (MOOPs), which are highly relevant problems in practice \cite{tian2021evolutionary}. In the optimization community, Multi-Objective Evolutionary Algorithms (MOEAs) are widely developed to solve MOOPs, through an iterative process of proposing solutions to the problem based on a set of previously proposed solutions.  
To optimize the performance of a MOEA, one needs to tune a set of parameter values. The process of tuning these parameters is often cumbersome and will lead to an (sub-)optimal set of values that is specifically tailored to the underlying problem. 
According to the ``No free lunch" theorem \cite{wolpert1997no}, 
there does not exist a set of parameter settings that are superior to other settings in all stages of optimization, or on all optimization problems. 
Besides the challenge of finding generalizable parameter settings in MOEA or evolutionary algorithms in general, another known challenge is their high computation time in converging to good solutions, especially for solving complex and high dimensional optimization problems. 

A way to circumvent the need to finetune the parameters of the MOEA is to implement Adaptive Parameter Control (APC) \cite{aleti2016systematic}. By adaptively setting the values dictating the behavior of the operators within the algorithm, APC is able to tailor the different operator settings to the different stages of optimization. Besides allowing for increased performance, it also eliminates the cumbersome task executing parameter optimization. As a MOEA operates using an iterative process of creating a population based upon the previous population, sequential decision making shows beneficial for this task. For this reason Deep Reinforcement Learning (DRL) shows most salient, 
as it considers the optimization performance attained by the algorithm until the current generation while it proposes new parameter settings.


In this paper, we implement APC using DRL on a MOEA. More specifically, we develop a framework that uses a Dueling Deep Q-Network (DDQN) agent to dynamically set the operators dictating the behavior of the Non-dominated Sorting Genetic Algorithm III (NSGA-III). The performance of the framework is tested on both a simple benchmark problem and a real-world Warehouse Design and Control Problem (WDCP), which concerns simultaneous optimization of warehouse layout design and the underlying control policies. 
By simultaneous optimization of its constituent sub-problems, solving the WDCP allows to take the (hierarchical) interrelations between the problems into account. This results in higher solution quality, albeit at an increased computational cost. Furthermore, we demonstrate the generalisability of the proposed approach by solving WDCP 
with the policies trained on a much simpler optimization problem, without the need for adjustments to the underlying method and without retraining of the DDQN agent.

The contributions of this work are summarized as follows:

\begin{itemize}
    \item We propose a framework for adaptive parameter control of a Multi-Objective Evolutionary Algorithm  that generalizes to different problems. To the best of our knowledge, we are the first to propose a DRL parameter control approach for a Multi-Objective Evolutionary Algorithm.

    \item We effectively train a Deep Reinforcement Learning policy for adaptive parameter control on a cheap-to-evaluate multi-objective optimization problem and deploy it to a real-world, computationally expensive, multi-objective combinatorial optimization problem. 
\end{itemize}

 

\section{Preliminaries and related work}\label{sec:RELATED}

\paragraph{Multi-Objective Optimization (MOO)} 
is a branch of problems involving two or more objective functions to be simultaneously optimized. 
MOO can be mathematically expressed as follows:
$$
\begin{array}{ll@{}ll}
\text{Minimize} \quad F(x)=\left\{f_{1}(x), f_{2}(x), \ldots, f_{m}(x)\right\} \\
\text{Subject \; to} \, \quad g(x) \leq 0
\end{array}
$$

where $x$ is the vector of design variables, $f_{i}(x)$ is the \textit{i}th objective function, and $g(x)$ is the constraint vector. The solution of MOO is generally expressed as a set of Pareto optima, representing the optimal trade-offs between the different objective values. Solutions present in this Pareto optima are Pareto optimum or non-dominated solutions, and a feasible solution $x*$ is Pareto optimal if there exists no feasible vector $x$ such that $f_{i}(x) \leq f_{i}\left(x^{\star}\right), i \in\{1,2, \ldots, m\}$ and $f_{i}(x)<f_{i}\left(x^{\star}\right)$, for at least one objective $i \in\{1,2, \ldots ., m\}$.

The multi-objective evolutionary algorithm (MOEA) has been demonstrated as an effective method for solving MOO problems, among which 
the Non-dominated Sorting Genetic Algorithm III (NSGA-III) is one of the most successful ones \cite{deb2013evolutionary}. MOEAs iteratively create a set of solutions, named a population of individuals, based on the well-performing solutions found in the previous population. The evolutionary process consists of three operators, the crossover operator, mutation operator and selection operator. The crossover operator takes two individuals, called parents, to create two new individuals, called childs, through recombination of the constituent parts of both parents. Thereafter the childs are taken through mutation, which randomly changes values to insert new values into the process. Finally, the parents and their childs are taken through the selection operator, which decides which individuals to remain and which to discard. Every cycle through all operators is called a generation, which are repeated until a set termination criterion is reached.

The DTLZ benchmark test suite is a widespread test suite for multi-objective problems with scalable fitness dimensions \cite{cheng2017benchmark}.  All problems in this test suite 
are continuous n-dimensional multi-objective problems. 

The Hypervolume Indicator (HV) is a frequently used performance measure for multi-objective optimization 
\cite{zitzler2004indicator}. The hypervolume is described as the volume of the space in the objective space dominated by the Pareto front approximation $Y_{N}$, which is the set of optimal solutions found, and bounded from above by a reference point $r \in \mathbb{R}^{m}$ such that for all solutions $y \in Y_{N}, y \le r$. The hypervolume indicator is computed as:
$$
H V\left(Y_{N} ; r\right)=\lambda_{m}\left(\bigcup_{y \in Y_{N}}[y, r]\right),
$$

\noindent in which $\lambda_{m}$ refers to the \textit{m}-dimensional Lebesgue measure for assigning a measure to the m-dimensional Euclidean space. It is defined by first calculating the Lebesgue outer measure, which implies the infimum of all values on the interval comprised by the values in the Pareto frontier. The values obtained through calculation of this infimum, show the dominated area and thus the hypervolume indicator value. A larger value of the hypervolume indicates a better found approximated Pareto front. By tracking the hypervolume indicator over consecutive generations, the convergence rate of the Pareto front can be analyzed.

\paragraph{Deep reinforcement learning for parameter control.}

Several existing works use DRL as APC in evolutionary computation to solve optimization problems. All these works consider the Differential Evolution (DE) algorithm. \citet{sharma2019deep} uses a Double Deep Q-Network (DQN) agent to aid the selection of different mutation operators within DE. The state representation focused on the locations and distribution of separate individuals in the solution space. \citet{tan2021differential} propose a somewhat similar approach using DQN, but their state representation focuses on the complexity of the solution space itself.  \citet{sun2021learning} implement a Policy Gradient method as the DRL algorithm. All these works use the CEC2013 and CEC2017 benchmarks to test the performance of their approaches. 
Our approach distinguishes itself in several aspects. First, we adopt the NSGA-III algorithm. The benefit of NSGA-III is the ability to maintain individual objective values throughout the process of multi-objective optimization. Doing this increases the utility of the approach for human decision-makers as they gain insight into the distribution of performance over the separate objective values. More importantly, with DRL as APC for NSG-III, the learned policy is highly generalizable. The agent is trained on a simpler problem with fast training time, and then it can be deployed to solve a much more complex problem. Through applying DRL as APC on the NSGA-III, we were able tackle a challenging, real-world optimization problem.

\section{Solution framework DRL-MOEA}\label{sec:SOLUTION}
The intended purpose of applying DRL is to increase the convergence speed without damaging the performance of the NSGA-III. Doing this will keep the benefit of using NSGA-III, namely the optimization performance, while mitigating the downside, namely the high computational cost. The benefit of using the proposed framework, instead of the regular NSGA-III, becomes especially apparent when the underlying problem increases in complexity. The additional computational cost of training an agent can be mitigated by training the agent on a more simplistic optimization problem before deploying it to a more complex problem. 


\subsection{DRL for APC}
We model the Adaptive Parameter Control (APC) problem for NSGA-III as a sequential decision problem. In the implementation of DRL in APC, it concerns learning policies of setting the parameter values dynamically during the iterations of the genetic algorithms to optimize the total reward. 


The implementation of DRL will be done using the offline, model-free Deep Q-Network (DQN). DQN learns to predict the values of being in a certain state, and in every state selects the action leading to the next state with the highest value. Taking these most ``profitable" actions will result in the creation of a so-called optimal behavioral policy, which tells the agent to take which action in which state. As evaluation time is the main concern of this research, training needs to happen using as little evaluations as possible. In this case, DQN has some beneficial characteristics. These characteristics concern sample efficiency and robustness, which are attained through implementation of experience replay and a frozen target network respectively. Experience replay concerns a buffer of interactions, frequently sampled to train the weights of the underlying neural network. The frozen target network, on the other hand, implies the agent using two neural networks. One network is used to dictate the behavior of the agent, where the other is used to predict the value of being in a certain state. The weights of the value network are repeatedly updated, using samples taken from the experience replay buffer. Only periodically, the learned weights will be copied onto the behavioral network, which dictates the behavior of the agent. This way the behavior of the agent remains stable, preventing it to fall into a positive feedback loop in which it chases an action returning a desired return.

The variation of the DQN used is called the Dueling DQN, or DDQN. The additional characteristic concerns a so-called Advantage function \cite{sewak2019deep}. This function returns an advantage value for all possible actions the agent can take, when in a certain state. These advantage values tell how much better it would be to take action $a_k$ in state $s$ over all other possible actions $a \in \mathcal{A}$ in state $s$. The advantage values together with the state values are used to calculate the Q-value(s), with 
$Q(s;a;\theta,\alpha,\beta) = V(s;\theta,\beta)+ (A(s, a;\theta,\alpha) - \frac{1}{|\mathcal{A}|} \sum_{a'}A(s;a';\theta;\alpha))$. 
The main benefit of using the advantage function besides the state values is to generalize learning across actions without imposing any change to the underlying reinforcement learning algorithm \cite{wang2016dueling}. Dueling DQN (DDQN) is beneficial if there could be cases where two actions have identical value, which is likely to happen given the environmental design the agent is deployed in.

\paragraph{State representation.}
The state representation concerns a description of the current state the agent is in. This representation will be taken through the behavioral network of the DDQN agent, which will return an action to be executed. 
Since the purpose of the proposed solution method is to be generalizable, all variables comprising the representation need to be made generic. The constructed representation focuses on three parts, namely, (1) the progress of the evolutionary process; (2) the performance and spread of the individuals comprising the  population; and (3) the performance and size of the found Pareto optimal set.

The evolutionary progress is measured using two variables, the current generation ($\mathcal{G}$) and the stagnation counter ($\mathcal{S}$). The current generations ($\mathcal{G}$) gives the agent insight into how much time remains to optimize, where the stagnation counter ($\mathcal{S}$) implies how many generations it did not produce any improvement. The stagnation counter is clipped, meaning that a maximal value is set. The stagnation counter value cannot exceed 10, which is a boundary that is only reached in extreme cases concluded from empirical evaluations. 

The population performance and spread is summarized in three values. The first consists of the average of all normalized objective values ($O_{mean}$), providing an indication of the overall performance of the population. The second value concerns the average of all normalized minimal values per objective value ($O_{min}$), focusing on the best obtained values found so far. The third value describes the average normalized standard deviations ($\sigma$) of all objectives. The reason the values are aggregated into a single value is to allow for generalization. By using aggregates, the agent can be applied to any optimization problem, independent of the number of objective values considered. 

The last part of the state representation describes the performance of the current approximation of the Pareto front. This performance is summarized in two metric, the hypervolume indicator ($\mathcal{H}$) and the size of the Pareto set ($PS$). The hypervolume indicator ($\mathcal{H}$) is the most used set-quality indicators for the assessment of multi-objective optimizers when the actual Pareto front is unknown \cite{guerreiro2020hypervolume}. 
The Pareto size ($PS$) is included to show the agent to what extend the maximum size of the front is reached. If the pareto size has not reached its maximum, there is a possibility to extend the Pareto optimal set without the need to discard another solution.


\paragraph{Action space.} 
The action space of the agent concerns the decisions it is able to make. With regard to NSGA-III, three values are considered to be included in this space. These values are summarized in \ref{tab:action_space} below, also showing the current values used in the initial implementation of the NSGA-III \cite{deb2013evolutionary} and describing the effect of high and low values.

\begin{table*}[ht!]
\centering
\footnotesize
\caption{
Action space Deep Reinforcement Learning (DRL) agent}
\begin{tabular}{l|lll}
\textbf{}                     & \textbf{$\eta_{SBX}$}                                                                & \textbf{$\eta_{PLM}$}                                                            & \textbf{$indpb_{PLM}$}                                                         \\ \hline
\textbf{Operator}       & Crossover                                                                                      & Mutation                                                                                 & Mutation                                                                             \\ \hline

\textbf{Description}          & \begin{tabular}[c]{@{}l@{}}Crossover \\ distribution parameter\end{tabular}             & \begin{tabular}[c]{@{}l@{}}Mutation \\ distribution parameter\end{tabular}         & \begin{tabular}[c]{@{}l@{}}(Independent)\\ mutation probability\end{tabular}  \\ \hline
\textbf{NSGA-III value}       & 30                                                                                      & 20                                                                                 & 0.01                                                                              \\ \hline
\textbf{Effect of high value} & \begin{tabular}[c]{@{}l@{}}Produce children \\ resembling their parents\end{tabular}    & \begin{tabular}[c]{@{}l@{}}Produce mutation \\ resembling its origin\end{tabular}    & \begin{tabular}[c]{@{}l@{}}Higher chance for \\ values to be mutated\end{tabular} \\ \hline
\textbf{Effect of low value}  & \begin{tabular}[c]{@{}l@{}}Produce children \\ dissimilar to their parents\end{tabular} & \begin{tabular}[c]{@{}l@{}}Produce mutation \\ dissimilar to its origin\end{tabular} & \begin{tabular}[c]{@{}l@{}}Lower chance for \\ values to be mutated\end{tabular} 
\end{tabular}\label{tab:action_space}
\end{table*}

The values described in \ref{tab:action_space} are solemnly considered. Actual construction of the action space relies on additional information gained from the process of Bayesian hyperparameter tuning, from which regions of well-performing hyperparameter settings can be extracted. 
The actual action space will be made discrete, having their values uniformly taken from the parametric ranges gathered from the hyperparameter tuning. The main reason the action space is made discrete is that small changes in the considered values do not impact behavior significantly. By setting discrete value pairs the selection of different action can be made significantly different, which we assume to be desirable for the implementation of the DDQN agent.

\paragraph{Reward function.} 
The intended purpose of implementing DRL as APC in the NSGA-III is to improve its convergence speed while maintaining solution quality. Hence, the implementation of the DDQN agent is not immediately on the problem, instead its environment consists of the underlying evolutionary algorithm. For this reason, the objective of the agent is different than the actual objectives of the given optimization problem. 
The DRL objective concerns fast convergence to a Pareto front while maintaining the quality of the found set of solutions. 
Scale sensitivity is of high importance when constructing a reward function for this purpose \cite{karafotias2015evaluating}. Within the NSGA-III, this implies the relative difference in attainable improvement comparing the first generations to the final generations. 
Inspired by \citet{huang2021operator}, we define our reward function as an episodic reward, calculated by the sum of all hypervolume indicator values over an entire optimization run:
$Reward_{gen} = \sum^{\left[1, ..., \mathcal{G}\right]}_{g=0} HV_{g}$, $ \forall  gen  \in \left[1, ..., \mathcal{G}\right]$. This metric includes both convergence speed and convergence quality, as an increase in either will increase the metric value. Using this metric will result in the agent optimizing its behavior towards converging in as few generations as possible to a set of solution having the highest solution quality possible.

\begin{figure*}[ht!]
    \centering
    \includegraphics[width=0.70\textwidth]{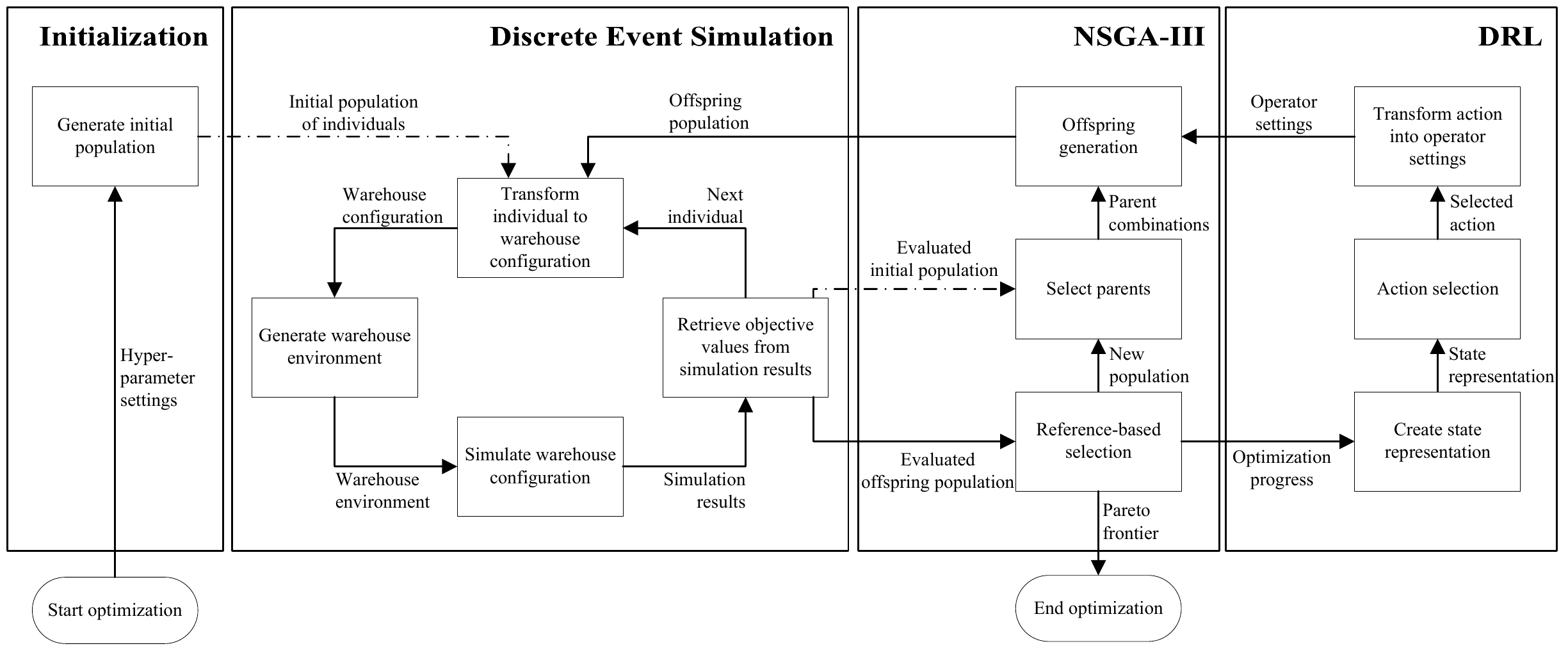}
    \caption{DRL-MOEA for WDCP.}
    \label{fig:Solution_Framework}
\end{figure*}

\subsection{Warehouse Design and Control Problem (WDCP)}\label{sec:CASE STUDY}

As stated earlier, the purpose of applying DRL is to increase the convergence speed without damaging the performance of the NSGA-III. Doing 
this will keep the benefit of using NSGA-III, namely the optimization performance, while mitigating
the downside, being the high computational cost. The benefit of using the proposed framework,
instead of the regular NSGA-III, becomes especially apparent when the underlying problem increases
in complexity. To validate this statement, the proposed approach will first be trained and tested on a simple problem after which it will be deployed on a complex real-world problem, being the Warehouse Design and Control Problem (WDCP), without retraining. For the
latter a simulation model is created, based on an actual warehouse of a large plastic manufacturer, to
evaluate the performance of the individuals proposed by the NSGA-III. A visual representation of the
entire solution framework for optimizing the WDCP, including the warehouse simulation, is shown in Figure \ref{fig:Solution_Framework}.

The studied WDCP consists of three sub-problems, namely the warehouse lay-out design problem, the resource allocation problem and the product allocation problem. The decision variables can be subdivided over four categories. The first two categories concern values dictating the Product Placement Algorithm (PPA). These values concern four ordinal values ($\mathcal{O}$), dictating the order of the rules within the PPA, and five parameter values ($\delta_{1}, \delta_{2}, \delta_{3}, \delta_{4}, \delta_{5}$), dictating the actual behavior of every single rule. The third category of decision variables concerns resource allocation and decides the number of resources per resource type ($\mathcal{Z}_{1} \hdots \mathcal{Z}_{n}$). Finally, the fourth category concerns the dimensions of storage locations. These dimensions can differ based on either width ($\mathcal{W}$), height ($\mathcal{H}$) or both. A formal definition of the number of storage locations having a given width/height combination is however trivial. An abstract representation of the amount of storage locations per hall ($\mathcal{X}$) can be described as $d^x_{w, h} \; \forall x \in \mathcal{X}, \forall w \in \mathcal{W}, \forall h \in \mathcal{H}$. 

The WDCP concerns multiple objectives: the tardiness of outbound trucks, the total cost of resources, and the number of unplaceable products. Upon arrival of an inbound truck ($\mathcal{K}_{inb}$) the truck is placed at one of the truck docks ($o_{k}$). After docking, all products are retrieved and placed throughout the warehouse in the locations determined using the PPA, dictated by the rules order ($\mathcal{O}$) and the parameter values ($\delta_{1}, \delta_{2}, \delta_{3}, \delta_{4}, \delta_{5}$). Transportation to these storage locations is executed using the available resources ($\mathcal{Z}_{1} \hdots \mathcal{Z}_{n}$). The number of storage locations of different dimensions in the different storage halls ($\mathcal{X}$) depends on $d^x_{w, h} \; \forall x \mathcal{X}, \forall w \in \mathcal{W}, \forall h \in \mathcal{H}$. As products can only be stored in storage areas equal or larger in either one or both of the dimensions, the possible storage locations of a given storage type ($\mathcal{S}$) for a given product can be defined as $d^{\mathcal{S}, x}_{p} \; \forall w \in \left[p_w, ..., \mathcal{W}^\mathcal{S}\right], \forall h \in \left[p_h, ..., \mathcal{H}^\mathcal{S}\right], \forall x \in \left[1, ..., \mathcal{X}^\mathcal{S}\right]$. If no available storage location can be found, the product is classified unplaceable, removed from the simulation and the average transportation time is used for the time a resource is occupied for placing the product in the warehouse.

The remaining two objective values concern the outbound process. Upon arrival of an outbound truck ($\mathcal{K}_{outb}$), it is placed at one of the available truck docks ($o_{k}$). Thereafter the requested products are retrieved one-by-one using the available resources ($\mathcal{Z}_{1} \hdots \mathcal{Z}_{n}$). The tardiness of these outbound trucks ($y_{k}$) is calculated by taking the difference between the departure time ($d_{k}$) and arrival time ($a_{k}$), multiplied with a penalty factor ($f_{k}$). This penalty takes a value of 0 for trucks departing within 30 minutes, 0.5 if trucks depart between 30 and 120 minutes and 1 for truck having a departure time exceeding 120 minutes. A mathematical formulation for this objective value is $y_{k} = \sum_{k \in \mathcal{K}_{outb}} f_{k} \cdot y_{k}$. The final objective value, total resource cost, is calculated as $\sum_{i=1, ... ,n} c_{i} \cdot \mathcal{Z}_{i}$, where $\mathcal{Z}_{i}$ describes the amount of resources for resource type i and $c_{i}$ concerns an aggregate of both investment and operating costs of that resources.


Optimization is executed using the NSGA-III algorithm. As previously explained, the NSGA-III proposes solution methods based upon previously evaluated solutions. A proposed solution consists of 96 values, including the rule order values ($\mathcal{O}$), the rule parameter values ($\alpha, \beta, \gamma, \delta, \upsilon$), the amount of resources per resource type ($\mathcal{Z}_{1} \hdots \mathcal{Z}_{n}$) and the amount of storage locations dedicated to a given set of dimensions per storage hall ($d^x_{w, h} \; \forall x \in \mathcal{X}, \forall w \in \mathcal{W}, \forall h \in \mathcal{H}$). These solutions are given to the Discrete Event Simulation (DES), which creates an environment based upon the given values. Thereafter the configuration is evaluated by running a simulation of 48 hours, of which 8 hours is a warm-up period to prevent a simulation bias due to starting with a warehouse that is not in equilibrium yet.

The NSGA-III repeats the generational cycle by continuously proposing solutions, combining and mutation solutions and proposing new ones. Within every generation the framework creates a state representation right after the selection process of the NSGA-III. This state representation is thereafter used by the DDQN agent to dynamically set the values dictating the behavior of the mutation operator. Through dynamic operator control, the algorithm gets tailored to the progress made in optimization so far. This process is repeated until a given termination criterion is reached, which is set to 200 generations.

\section{Experimental results}\label{sec:RESULTS}
All experiments were run on a single computer containing 12 Intel I7-8750H CPUs, having 15.5 GB RAM at its disposal. The results regarding optimization performance  show both an average and standard deviation of a set number of independent optimization trajectories. The amount of independent optimization runs used is dependent on the computation cost incurred and varies between the simple and complex problem. 
We compare the proposed DRL-MOEA approach with (1) the NSGA-III with optimized hyperparameter settings (optimized NSGA-III), and (2) an untrained (random) implementation of DRL-MOEA, which is the NSGA-III with adaptive parameter control (APC) where parameters are randomly selected, in contrast to those being learned in DRL-MOEA. 
The parameter settings used for all experiments can be found
in the provided Supplementary Material. 




\begin{figure}[ht!]
    \centering
    \includegraphics[width=0.42\textwidth]{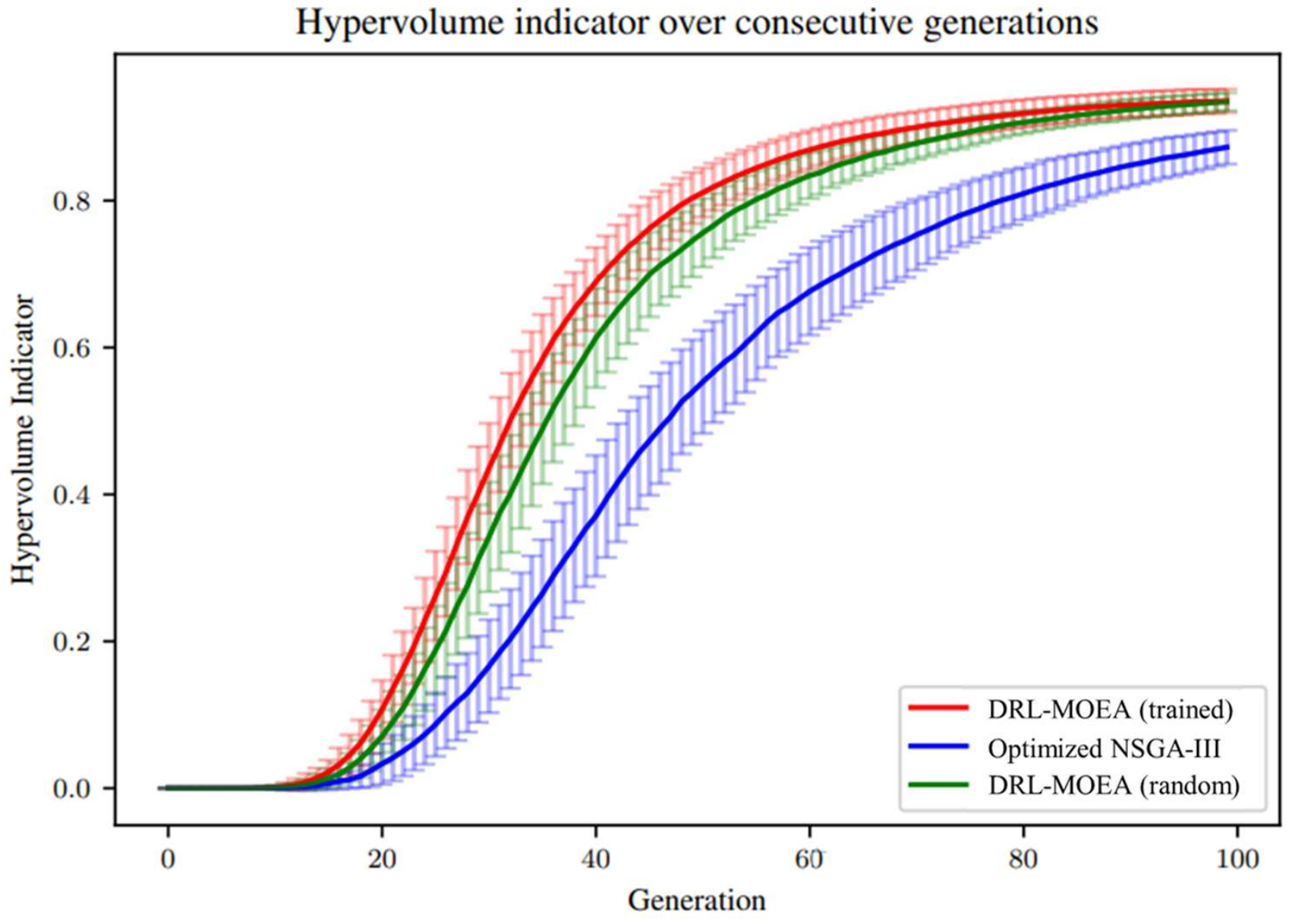}
    \caption{Performance benchmark DRL agent on the DTLZ2 benchmark}
    \label{fig:Agent_DTLZ2_Benchmark}
\end{figure}

\subsection{Performance on the DTLZ2 benchmark}\label{subsec:PERF_DTLZ2}
First, the agent is trained and tested using simple problems from the DTLZ benchmark test suite. Since results are similar, we report here only the performance on the DTLZ2 benchmark function. After 4000 episodes of 200 generations each, the performance of the learned policy is evaluated. Due to the relatively low computation cost, 500 independent optimization trajectories are used to evaluate the performance of the different models. To emphasize the effect on convergence speed, the evaluation of performance will focus on the first 100 generations. The remaining 100 generations will show an almost linear line, indicating the convergence to an optimal set of solutions to the problem. The attained performances of the different models is shown in \ref{fig:Agent_DTLZ2_Benchmark}.


 \begin{figure*}[ht!]
    \centering
    \includegraphics[width=0.70\textwidth]{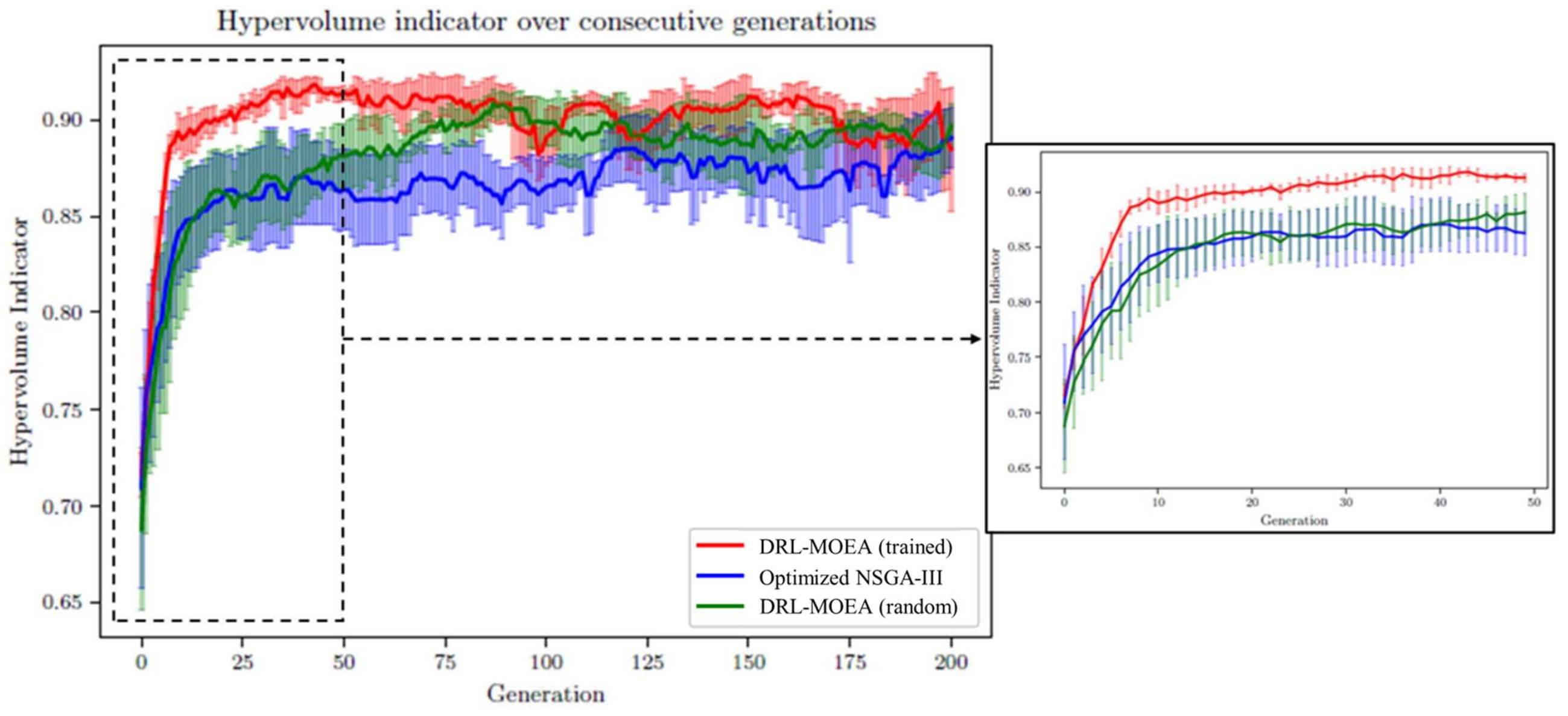}
    \caption{Performance benchmark DRL agent on the WDCP}
    \label{fig:Performance_WDCP}
\end{figure*}

Interesting to see is that both methods that apply APC, namely DRL\_MOEA (trained) and DRL\_MOEA (random), show an increase in optimization performance with regards to convergence speed. A possible explanation for the increased performance of the DRL\_MOEA (random) can be found in both the use of a small population size of 20 individuals and the structural design of the NSGA-III. Using a small population size applying radical mutation can be beneficial to escape local optima. Under the expectation that the DRL\_MOEA (random) will decide uniformly which action to take, 50\% of its actions will show an inclination towards radical mutation. Additionally, the selection operator within the NSGA-III diminishes the downside of radical mutation, as individuals with poor performance will be removed from the population. But still, the learned DRL\_MOEA outperforms the random DRL\_MOEA, which indicates the added benefit of learning a policy to apply mutation within the NSGA-III. 

\subsection{Apply learned policy to solve WDCP}\label{subsec:PERF_WDCP}
After learning an optimal policy on the DTLZ2 benchmark, the agent will be applied on the more complex problem, namely the Warehouse Design and Control Problem (WDCP). Solving the WDCP is done on an actual real-world use case, for which the warehouse of a large plastic manufacturer is replicated. Due to the high dimensionality and high level of detail of this simulation, the computation cost for a single evaluation is around 2-4 seconds. For this reason only five independent optimization trajectories are used to evaluate the three models. 
The results obtained are visualized in \ref{fig:Performance_WDCP}.


\begin{figure*}[ht!]
    \centering
    \includegraphics[width=0.9\textwidth]{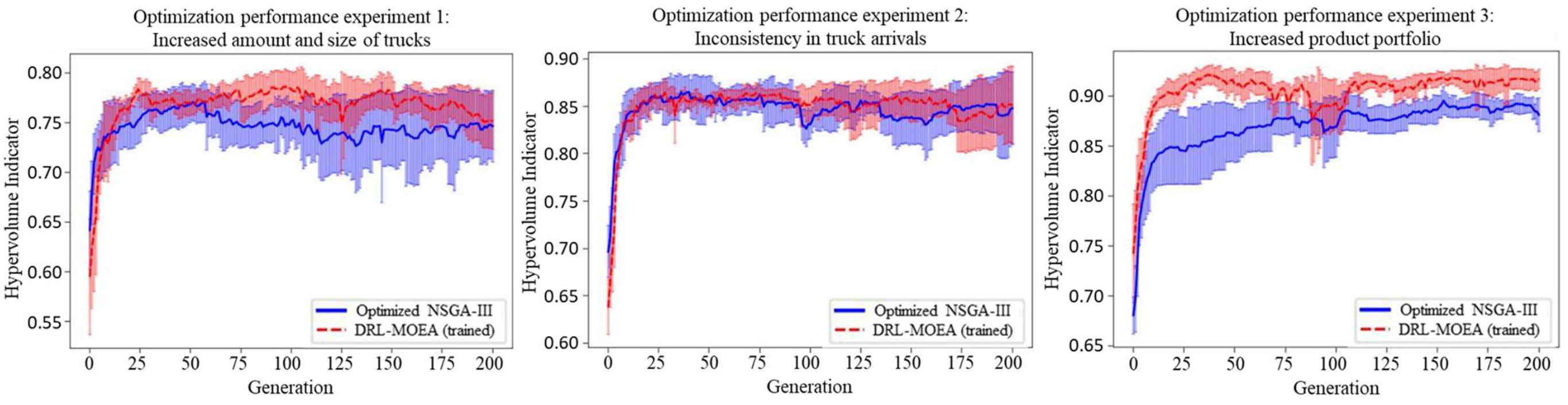}
    \caption{Sensitivity analysis of the learned agent on the WDCP}
    \label{fig:SA_WDCP}
\end{figure*}

Figure \ref{fig:Performance_WDCP} shows the performance of the learned agent (i.e. DRL-MOEA (trained)) is superior to the performance of both the DRL\_MOEA (random) and optimized NSGA-III, even though the agent was trained on a different problem, namely DTLZ2. The learned DRL\_MOEA (random) attained a maximum hypervolume indicator of 0.92, in comparison to 0.91 of the random DRL\_MOEA and 0.89 of the optimized NSGA-III.  
Until generation 50 the increased performance is significant, as emphasized in the cut-out. 
After generation 50, the NSGA-III with a random action selecting agent implementation of APC shows to gain performance, which might indicate it breaking free from a local optimum due to the previously mentioned probability of an inclination towards radical mutation. Besides convergence speed, also the volatility over the five consecutive optimization trajectories decreased using the learned agent.

To further evaluate the robustness of the learned agent, an additional sensitivity analysis is conducted. In this analysis three scenarios are tested, which all focus on one or more of the constituent sub-problems of the WDCP. The first analysis increases the amount and size of incoming trucks, increasing overall pressure on the warehouse. The second analysis removes the consistency of truck arrivals, increasing the complexity of the resource allocation problem. The third analysis increases the product portfolio, doubling the amount of distinct products, resulting in higher complexity of the product allocation problem. The results of all three analysis are shown in Fig. \ref{fig:SA_WDCP}, showing the difference in performance of using a learned agent or not. As with the initial analysis, the performances are calculated over five independent optimization trajectories. 
In two of the three analysis, the learned agent as APC in the NSGA-III has led to increased performance. The maximum obtained hypervolume indicator for the learned agent and the optimized NSGA-III without an agent are 0.79 and 0.77 for the first experiment, 0.86 and 0.86 for the second experiment and 0.92 and 0.90 for the third experiment respectively. For the second experiment it is expected that the performance attained by the optimized NSGA-III without APC is already the optimal performance attainable, thus no improvement was possible by implementing APC using the learned agent. Besides, the volatility between the five independent optimization trajectories decreased by using the learned agent in all three scenarios, indicating a more stable optimization process. 

\subsection{Policy analysis}\label{subsec:POLICY_EVAL}
Both on the DTLZ2 benchmark as well as the WDCP the learned agent outperforms the benchmarked methods. The fact that the learned agent is able to retain superior performance shows that the learned behavior is generalizable, as no additional training is executed for the WDCP. To gain further insight into the behavior of the learned agent, the action selection policy for both the DTLZ2 benchmark as well as the WDCP is further analyzed. The action space of the agent consists of 15 different combinations of mutation intensity (eta) and probability (indpb), where lower values indicate conservative mutation and higher values indicate more radical mutation. 

Besides gaining insight into the action selection behavior of the agent, this analysis can also be used to validate the need to use DRL in the first place. If the behavior for both problems shows to be identical, the agent possibly can be replaced by a set of rules deduced from the agent. If the behavior shows significantly different, it indicates the need for sequential decision and thus to use DRL.

\begin{figure}[ht!]
    \begin{center}
        \resizebox{0.5\textwidth}{!}{\input{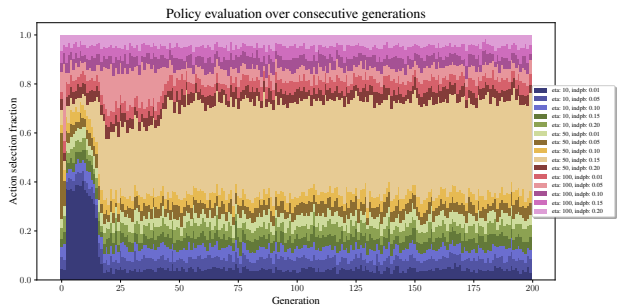}}
    \end{center}
    \caption{Policy visualization for the DTLZ2 benchmark problem over consecutive generations}
    \label{fig:DTLZ_Policy}
\end{figure}

Analysis of the behavioral policy for the DTLZ2 benchmark is based upon the actions taken in all 500 independent optimization trajectories. The fraction of action selection in every generation is shown in \ref{fig:DTLZ_Policy}. The agent behavior indicates that the agent starts with more conservative mutation until generation 20, after which the inclination towards more radical mutation increases between generation 20 and 50. The initial preference for conservative mutation is however counter intuitive. One would expect that such algorithms tend to first explore the search space after which they exploit the gained knowledge in exploration to improve. The behavior shown in \ref{fig:DTLZ_Policy} indicates the opposite, as the agent first inclines towards exploitation after which it gradually moves towards more exploratory behavior.

\begin{figure}[ht!]
    \begin{center}
        \resizebox{0.5\textwidth}{!}{\input{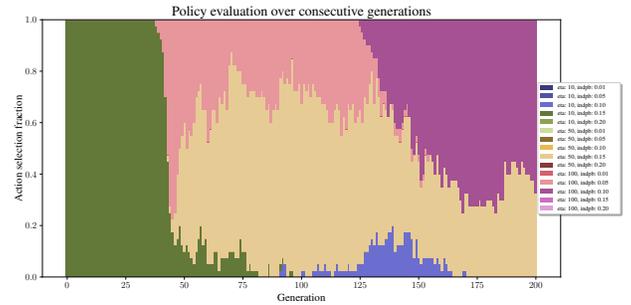}}
    \end{center}
    \caption{Policy visualization for the WDCP over consecutive generations)}
    \label{fig:Policy_WDCP}
\end{figure}

We analyze the behavior of the learned agent on the WDCP. We ran 40 independent optimization trajectories using the learned agent. The decrease in trajectories is due to the increased computation cost as a result of the simulation used in the WDCP. Figure \ref{fig:Policy_WDCP} shows the fraction of actions selected over consecutive generation for 40 independent optimization trajectories. Concluding from the action selection behavior of the learned agent in Figure \ref{fig:Policy_WDCP} an identical shift from conservative to radical mutation can be seen. However, the increased complexity of the underlying problem resulted in the agent being more decisive in its action selection. Compared to the DTLZ2 benchmark, in which all actions were selected at least once, only five different actions are selected by the agent on the WDCP. Also, the agent tends toward a higher intensity of mutation as opposed to the DTLZ2 behavior. This might be due to the WDCP being a non-convex optimization problem, which increases the need for radical mutation to escape the local optima in the search space.

\section{Conclusions}\label{sec:CONCLUSIONS}
Through this research, a generalizable solution framework is proposed that integrates the NSGA-III with APC, implemented using a Dueling DQN agent. The proposed framework initially shows increased performance on the DTLZ2 benchmark, which is the problem used throughout training of the agent. Thereafter the framework is applied to a complex real-world problem. 
The performance attained by the framework shows to be superior to that of the optimized NSGA-III without APC. The fact that the performance was retained when the underlying problem was changed and the model was not retrained, shows that the model is generalizable. Through a sensitivity analysis, increasing the complexity of the constituent sub-problems of the WDCP, this performance increase is maintained, which indicates the robustness of the framework. A difference in behavioral policy of the agent between the DTLZ2 benchmark and the WDCP shows the inability to replace the DRL agent with a simplistic set of rules. This validates the need to use DRL as a sequential decision making framework.

\bibliography{references}


\end{document}